\documentclass{article}
\usepackage[final]{corl_2018} 

\usepackage[utf8]{inputenc} 
\usepackage[T1]{fontenc}    
\usepackage{hyperref}       
\usepackage{url}            
\usepackage{booktabs}       
\usepackage{amsfonts}       
\usepackage{nicefrac}       
\usepackage{microtype}      
\usepackage{cleveref}  

\usepackage{graphicx} 
\usepackage{subfig}  
   \graphicspath{{figs/}}
   \DeclareGraphicsExtensions{.pdf,.jpg,.png,.eps}

\usepackage{siunitx}
\usepackage{multirow}

\usepackage{floatrow}
\newfloatcommand{capbtabbox}{table}[][\FBwidth]

\DeclareRobustCommand*\pct{\scalebox{.9}{\%}}

\usepackage{amsmath,amssymb}
\usepackage{xspace}

\newcolumntype{L}{S[table-format=2.1]@{\hskip 10pt}}
\newcolumntype{R}{S[table-format=2.1]}
\newcolumntype{?}{!{\vrule}}
\newcommand{\myparagraph}[1]{\textbf{#1}~}

\newcommand{\dense}{\thickmuskip=2mu}

\renewcommand{\eqref}[1]{(\ref{#1})}
\newcommand{\figref}[1]{Figure~\ref{#1}}

\newcommand{\ie}{\textrm{i.e.}}
\newcommand{\eg}{\textrm{e.g.}}
\newcommand{\etc}{\textrm{etc.}}
\newcommand{\etal}{\textrm{et~al.}}

\newcommand{\map}{\ensuremath{\mathbb{M}}\xspace}
\newcommand{\lik}{\ensuremath{f}\xspace}

\newcommand{\threed}{\mbox{3-D}\xspace}
\newcommand{\twod}{\mbox{2-D}\xspace}

\usepackage{color}
\definecolor{fullred}{rgb}{0.95,.0,.1}
\newcounter{cmt}

\title{
Particle Filter Networks with Application \\ 
to Visual Localization
}

\author{
  Peter Karkus$^{1,2}$
  \qquad
  David Hsu$^{1,2}$
  \qquad
  Wee Sun Lee$^2$  \vspace{0.25cm}\\
  $^1$NUS Graduate School for Integrative Sciences and Engineering  \\
  $^2$School of Computing\\
  National University of Singapore\\
  \texttt{\{karkus, dyhsu, leews\}@comp.nus.edu.sg} \\
}

\begin{document}
\maketitle


\begin{abstract}
  Particle filtering is a powerful approach to sequential state estimation and
  finds application in many domains, including robot localization, object
  tracking, \etc{} 
  To apply particle filtering in practice, a critical challenge is to
  construct probabilistic system models, especially for systems with complex
  dynamics or rich sensory inputs such as camera images.  This paper
  introduces the Particle Filter Network~(PF-net), which encodes both a system
  \emph{model} and a particle filter \emph{algorithm} in a single neural
  network.  The PF-net is fully differentiable and trained end-to-end from
  data. Instead of learning a generic system model, it learns a model
  optimized for the particle filter algorithm.
  We apply the PF-net to a visual localization task, in which a robot must
  localize in a rich 3-D world, using only a schematic 2-D floor
  map. 
  In simulation experiments, PF-net consistently outperforms alternative
  learning architectures, as well as a traditional model-based method, under a
  variety of sensor inputs. Further, PF-net generalizes well to new, unseen
  environments.
\end{abstract}


\section{Introduction}
\begin{figure*}[!t]
  \centering
   \sidesubfloat[][]{ \hspace{-0.15cm}\includegraphics[height=2.5cm]{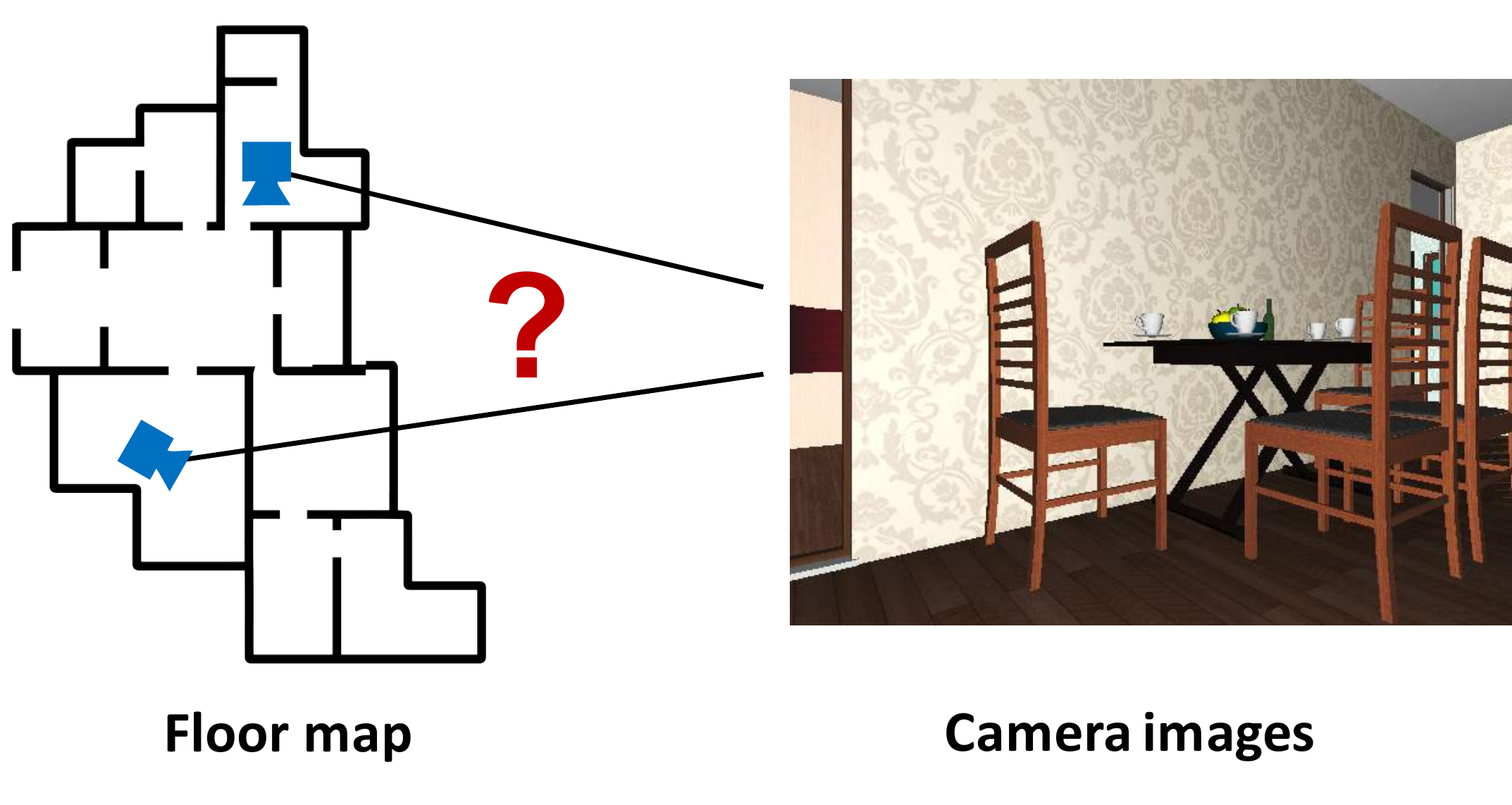}\label{fig:visual_localization} \hspace{0.14in}} 
   \sidesubfloat[][]{ \hspace{-0.15cm}\includegraphics[height=2.5cm]{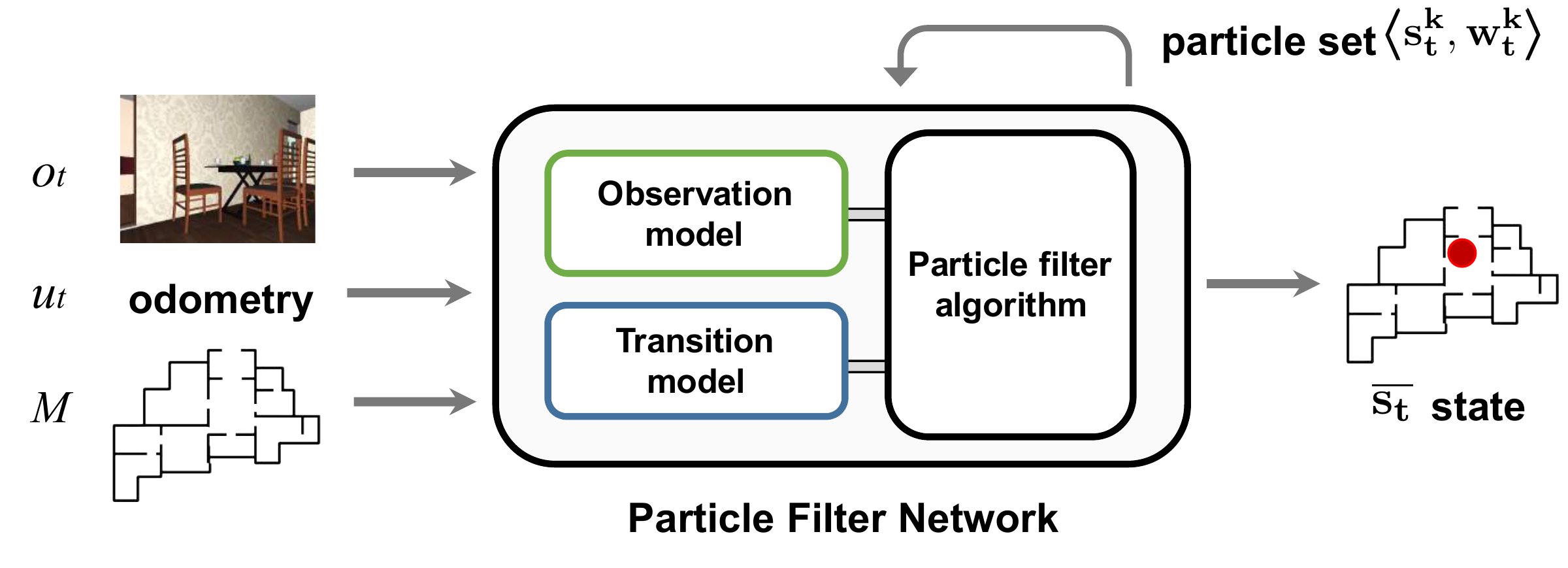}\label{fig:architecture} }
   \caption{ (a) Robot visual localization in a \threed environment,
     using only a schematic \twod floor map.  The robot must match rich
     \threed visual features with crude \twod geometric features from the map.
     It must also ignore objects not in the map, \eg, furniture.  
     (b)~PF-net 
     encodes both a learned probabilistic system model and the particle filter
     algorithm in a single neural network. It trains the model end-to-end in the
     context of the particle filter  algorithm, resulting in improved
     performance.
	\vspace{-6pt}
   }
\label{fig:fig1_combined}
\vskip -.07in
\end{figure*}

Particle filtering, also known as the sequential Monte-Carlo method, is a
powerful approach to sequential state estimation~\cite{doucet2001introduction}.
Particle filters are used extensively in
robotics, computer vision, physics, 
econometrics, \etc~\cite{thrun2002particle, blake1997condensation,
  ristic2003beyond, del2006sequential, liu2008monte, doucet2001sequential}, 
and are critical for robotic tasks such as
localization~\cite{thrun2001robust}, SLAM~\cite{montemerlo2002fastslam}, and
planning under partial observability~\cite{ye2017despot}.
To apply particle filters in practice,
a major challenge is 
to construct probabilistic system models or learn them from
data~\cite{shani2005model, boots2011closing, getoor2002learning}.  Consider,
for example, robot localization with an onboard 
camera~(\figref{fig:visual_localization}).
The observation model
is a probability distribution over all possible camera images, conditioned on
a continuous robot state and an environment map.  Learning such a model is
challenging, because of the enormous observation space and the lack of sufficient
labeled data. An emerging line of research circumvents the difficulty of traditional model 
learning: it embeds an algorithm into a deep neural network
and then performs end-to-end  learning to train  a model optimized for the specific
algorithm~\cite{tamar2016value, 
karkus2017qmdp, oh2017value, farquhar2017treeqn, amos2017optnet}.

In this direction, we introduce the Particle Filter Network~(PF-net), a
recurrent neural network~(RNN) with differentiable algorithm prior
for sequential state estimation. 
A PF-net encodes learnable probabilistic state-transition and observation
\emph{models} together with the particle filter \emph{algorithm} in a single
neural network~(\figref{fig:architecture}).  
It is fully differentiable and trained end-to-end from data.
PF-net tackles the key challenges of learning complex probabilistic system
models.  Neural networks are capable of representing complex models over large
spaces, \eg, observation models over images. Further, the network
representation unites the model and the algorithm and thus allows 
training end-to-end. As a result, PF-net learns system models optimized 
for a specific algorithm, in this case, particle filtering,
instead of learning generic system models.
The models may learn only the features relevant for state estimation,
thus reducing the complexity of learning.

\floatsetup[figure]{style=plain,subcapbesideposition=top}

We apply PF-net to robot visual localization, which is of great interest to
mobile robotics.  A robot navigates in a previously unseen environment and
does not know its own precise location.  It must localize in a visually rich 3-D world,
given only a schematic 2-D floor map and observations from onboard
sensors~(\figref{fig:visual_localization}).  While particle filtering is the
standard approach for
LIDAR~\cite{thrun2001robust}, 
we consider visual sensors, \eg, cameras.  Now the probabilistic observation
model must match rich \threed visual features from camera images to crude
\twod geometric features from the map.  Further, the camera images may
contain various objects not in the map, \eg, furniture.  This task exhibits
key difficulties of state estimation from ambiguous, partial
observations.  A standard model-based approach would construct an observation
model as a probability distribution of images conditioned on the floor map 
and robot pose. 
This is difficult, because of the enormous observation space, \ie, the space
of all possible images showing various floor layouts, 
furniture configurations, \etc~In contrast, PF-net trains a
model end-to-end and learns only  features relevant to the localization task.

This paper makes two contributions. First, we encode a particle filter
algorithm in a neural network to learn models for sequential state estimation
end-to-end.  Second, we apply PF-net to visual localization and present a
network architecture for matching rich visual features of a 3-D world with a
schematic 2-D floor map.  Simulation experiments on the House3D data
set~\cite{wu2018building} show that the learned  PF-net is effective for visual
localization in new, unseen environments populated with furniture.  Through
end-to-end training, it also outperforms  a conventional
model-based method; it fuses information from multiple sensors, in particular,
RGB and depth cameras; and it naturally integrates semantic 
information for localization, such as map labels for doors and room types.

\section{Background}

\subsection{Related work}

The idea of differentiable algorithm priors, \ie, embedding algorithms into a deep neural
network, has been gaining attention recently.
It has led to promising results for graph search~\cite{oh2017value, farquhar2017treeqn,
  guez2018learning}, path integral optimal control~\cite{OkaRig17},
quadratic optimization~\cite{amos2017optnet, donti2017task},
and decision-making in fully observable environments~\cite{tamar2016value} and
partially observable environments~\cite{karkus2017qmdp, shankar2016reinforcement}.

The general idea, when  applied  to probabilistic state estimation, has led
to, \eg, Kalman filter network~\cite{haarnoja2016backprop} and histogram
filter network~\cite{jonschkowski2016}. However, Kalman filtering assumes that the
underlying state distribution is or can be well approximated as a unimodal
Gaussian. Histogram filtering  assumes discrete state spaces and has difficulty
in scaling up to high-dimensional state spaces because of the ``curse of
dimensionality''. 
To tackle arbitrary distributions and very large discrete or continuous state spaces, 
one possibility is particle filtering.
Concurrent to our work, Jonschkowski~\etal{} have been independently
working on the idea of differentiable particle
filtering~\cite{jonschkowski2018differentiable}. The work is closely related, and
we want to highlight several important differences.  
First, we propose a differentiable approximation of resampling, a crucial step
for many particle filter algorithms.   
Next, we apply PF-net to visual localization in \emph{new, unseen} environments, 
after learning. While the concurrent work also deals with localization, it
does so in a  fixed environment.
Finally, our observation model for visual localization matches rich \threed
visual feature with a schematic \twod floor map, ignores objects not
in the map, and fuses information from multiple sources. 
Neural networks have been used with particle filters 
in variational learning as well.
Unlike the PF-net, such networks aim to parameterize a family of generative distributions
over observations~\cite{naesseth2017variational, maddison2017filtering,
  le2018auto, gu2015neural}, thus making them unsuitable for large, complex
observation spaces, such as the space of camera images and floor maps.

Particle filter methods, \eg, Monte-Carlo
localization~\cite{thrun2001robust}, are standard solutions to mobile robot
localization. 
Many such methods assume a LIDAR sensor mounted on the robot and rely on  
handcrafted simple analytic observation models~\cite{thrun2005probabilistic}. 
While there have been attempts to incorporate monocular or depth 
cameras~\cite{dellaert1999using, elinas2005sigmamcl, coltin2013multi, mendez2017sedar},
constructing probabilistic observation models for them  remains a challenge.  
PF-net learns effective system models through end-to-end training, without 
direct supervision on model components. 

\subsection{Particle filter algorithm}
Particle filters periodically approximate the posterior distribution over states
after an observation is received, \ie{}, they maintain a belief over states, $b(s)$.
The belief is approximated by a set of \emph{particles}, \ie{}, weighted samples
from the probability distribution, 
\begin{equation}\label{eq:particles}
b_t(s) \approx \langle s_t^k, w_t^k \rangle_{k=1:K},
\end{equation} 
where $\dense \sum_k w_k = 1$, $K$ is the number of particles, 
$s_k$ is the particle state, $w_k$ is the particle weight, and $t$ denotes time.
Importantly, the particle set can approximate arbitrary distributions, 
\eg, continuous, multimodal, non-Gaussian distributions. 
The state estimate can be computed by the weighted mean,
$\textstyle{\overline{s}_t = \sum_{k}{w_t^k s_t^k}}$.
The particles are periodically updated in a Bayesian manner.
First, the particle states are updated by sampling 
from a probabilistic transition model, 
\begin{equation}\label{eq:T}
s_t^k \thicksim T(s_t | u_t, s_{t-1}^k),
\end{equation}
where the transition model, $T$, defines the probability of a state, $s_t$,
given a previous state, $s_{t-1}^k$, and the last action, $u_t$. 
In the case of robot localization $u_t$ is the 
odometry input.
Second, the particle weights are updated. The likelihood, $\lik_t^k$, 
is computed for each particle,
\begin{equation}\label{eq:Z}
\lik_t^k = Z(o_t | s_t^k; \map),
\end{equation}
where the observation model, $Z$, defines the conditional probability of an observation, 
$o_t$, given a state and the \twod floor map, $\map$. 
Particle weights are updated according to the likelihoods,
\begin{equation}\label{eq:weight_update}
w_t^k = \eta \lik_t^k w_{t-1}^k,
\end{equation}
where  $\eta^{-1} = \sum_{j=1:K}{\lik_t^j w_{t-1}^j}$ is a normalization factor.

One common issue is particle degeneracy, \ie, when most particles have
near-zero weight. The issue can be addressed by \emph{resampling}
particles. New particles are sampled from the current set with repetition, 
where a particle is chosen with a probability proportionate to its weight,
\begin{equation}\label{eq:resample}
p(k)=w_t^k.
\end{equation}
The weights are updated according to a uniform distribution,
\begin{equation}\label{eq:reweight}
w_t^{\prime k} = 1/K.
\end{equation}
The new particle set approximates the same distribution, but devotes its
representation power to the important regions of the belief space. Note that 
the new set may contain repeated particles, but they diverge 
after stochastic transition updates.

\section{Particle Filter Network}

\begin{figure}[!t]
\begin{floatrow}
\ffigbox[.38\textwidth]{%
    \centering \hspace{-0.0cm}\includegraphics[height=2.35cm]{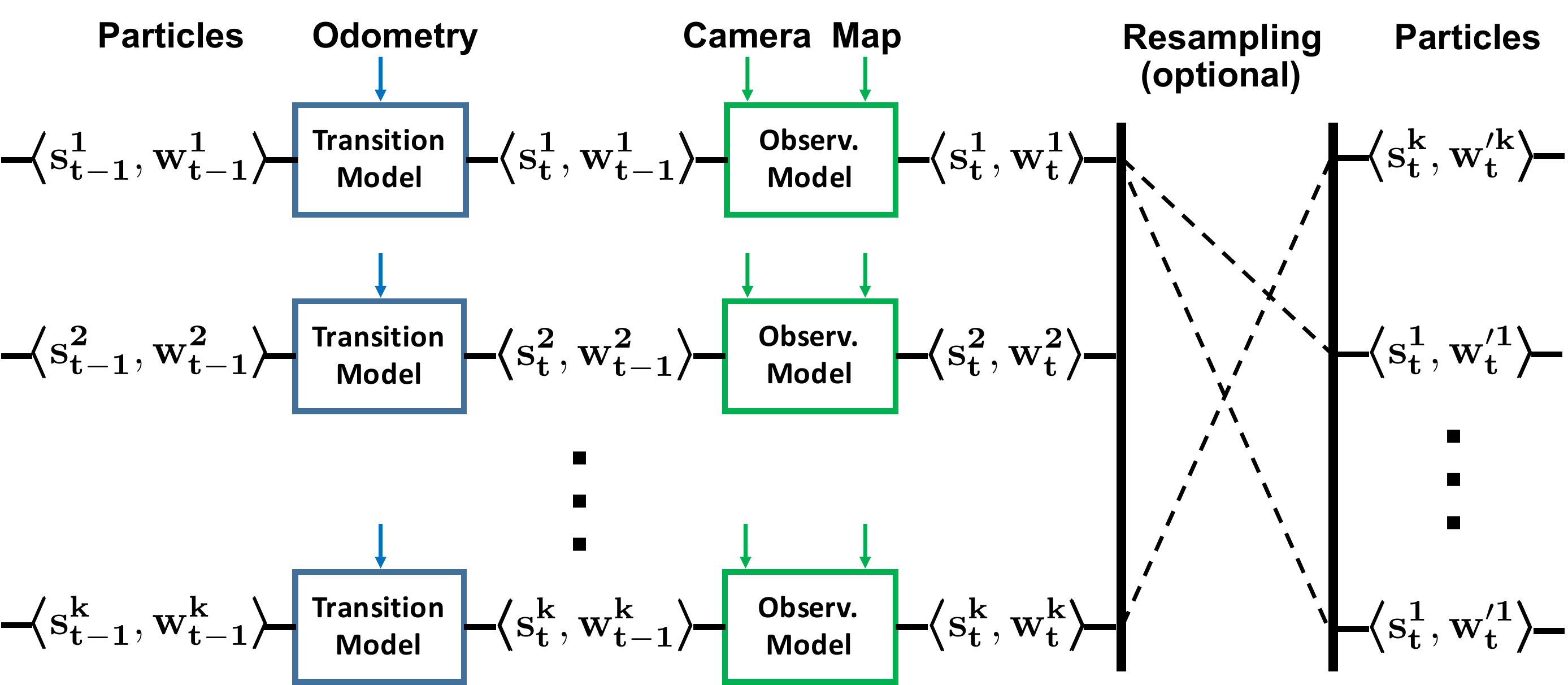}
}{\caption{PF-net as a computation graph. 
   The state-transition and observation models are captured in network
   weights, which  are shared across the particles.
\vspace{-6pt}%
}\label{fig:particle_update}
}
\quad
\ffigbox[.56\textwidth]{%
\centering \hspace{-0.0cm}\includegraphics[height=1.7cm]{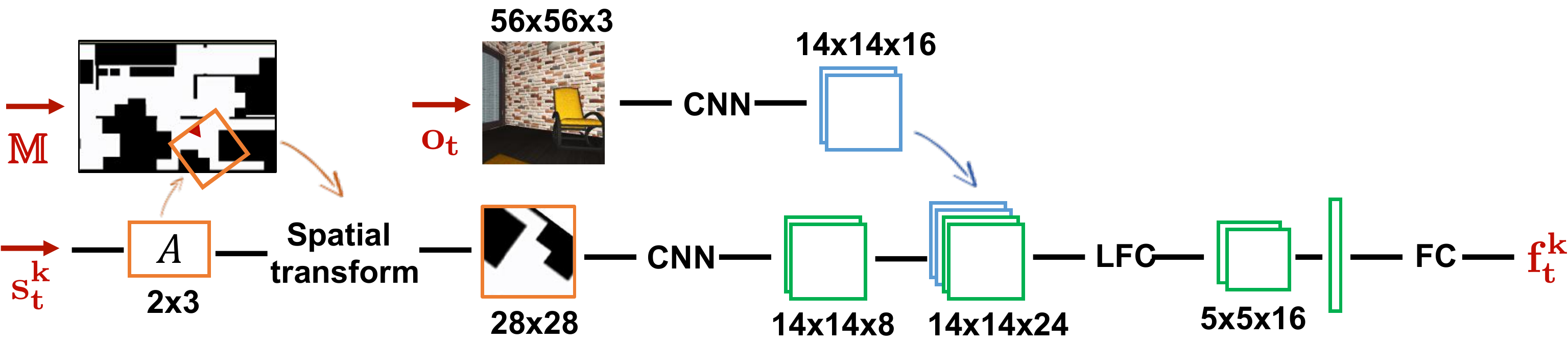}
}{\caption{The PF-net observation model.
 The inputs are floor map~$\map$, observation~$o_t$,
  and particle state~$s_t^k$. The output is particle likelihood~$\lik_t^k$. 
  CNN, LFC and FC are convolutional, locally fully-connected, and fully
  connected network components, respectively.
\vspace{-6pt}%
} \label{fig:obsmodel}
}
\end{floatrow}
\end{figure}

The Particle Filter Network~(PF-net) encodes learnable transition and 
observation models, together with the particle filter algorithm, in a 
single neural network~(\figref{fig:architecture}). 
PF-net is a RNN with 
differentiable algorithm prior, that is, structure specific
to sequential state estimation. 
The differentiable algorithm prior in PF-net is particle filtering: 
the particle representation of beliefs, and Bayesian updates for 
transitions and observations. 
Compared to generic architectures, 
such as LSTM~\cite{hochreiter1997long}, 
these priors allow much more efficient learning. 

The key idea 
underlying our approach is the unified representation of a
learned model and an inference algorithm.
The model is a neural network, \ie, a computation
graph with trainable parameters. The inference algorithm 
is a differentiable program, \ie, a computation graph with
 differentiable operations. Both the model and the algorithm 
can be encoded in the same computation graph.
What is the benefit?
Unlike conventional model learning methods, PF-net can learn a
model end-to-end, backpropagating gradients through the 
inference algorithm. The model is now optimized for
a specific inference algorithm and a specific task. As a result, the 
model may not need to capture complex conditional 
probability distributions, instead, it may learn only the features 
relevant to the task.

Specifically, PF-net encodes the particle filtering steps,  
{\eqref{eq:particles}-\eqref{eq:reweight}}, in a
computation graph~(\figref{fig:particle_update}).
The transition and observation models, \eqref{eq:T} and~\eqref{eq:Z}, are 
trainable neural networks with appropriate structure. 
Learned network weights are shared across particles.
The rest of the computation graph is not learned, but rather,
it implements the operations~{\eqref{eq:particles}-\eqref{eq:reweight}}.
Importantly, these operations must be 
differentiable to allow backpropagation. This is an issue for
sampling from a learned distribution in~\eqref{eq:T}, 
and resampling particles in~{\eqref{eq:resample}-\eqref{eq:reweight}}.

The sampling operation~\eqref{eq:T} is not differentiable, but it can be
easily expressed in a differentiable manner using the
``reparameterization trick''~\cite{kingma2013auto, doersch2016tutorial}. 
The trick is to take a noise vector as input, and express the 
desired distribution as a deterministic, differentiable function of this
input. The function may have learnable parameters,
\eg, the mean and variance of a Gaussian. 
Particle resampling poses a different issue: new particle weights 
are set to constant in~\eqref{eq:resample}, which produces zero gradients. 
We address the issue by introducing \emph{soft-resampling}, 
a differentiable approximation based on importance sampling. 
Instead of sampling particles from the desired distribution $p(k)$, 
we sample from $q(k)$, a combination of $p(k)$ and a uniform distribution,
\begin{equation}\label{eq:softresample}
q(k) = \alpha w_t^k + (1-\alpha) 1/K,
\end{equation}
where $\alpha$ is a trade-off parameter.
The new weights are computed by the importance sampling formula, 
\begin{equation}\label{eq:softreweight}  
w_t^{\prime k} = \frac{p(k)}{q(k)} = \frac{w_t^{k}}{\alpha w_t^k + (1-\alpha) 1/K}.
\end{equation} 

This operation has non-zero gradient when $\alpha \neq 1$. Soft-resampling 
trades off the desired sampling distribution~($\alpha=1$) with the uniform 
sampling distribution~($\alpha=0$). It provides non-zero gradients 
by maintaining the dependency on previous particle weights.
An alternative to soft-resampling is to simply carry over
particles to the next step, without resampling them. We found this 
to be a good strategy when training in a low uncertainty setting, \ie,
when most particles remain close to the underlying true states. 
Soft-resampling worked better under high uncertainty, where 
most particles would deviate far from the true states.

We have now introduced the PF-net architecture in a general setting.
When applying PF-net to a particular task, we must choose the 
representation of states, and the network architecture for $T$ and $Z$. 
Note that we may use different number of particles
during training and during evaluation.

\section{Visual localization}\label{sec:visual_localization}

We apply PF-net to visual localization~(\figref{fig:fig1_combined}). 
A robot navigates in an indoor environment 
it has not seen before. The robot is uncertain of its location.
It has an onboard camera, odometry, and it receives a 
schematic \twod floor map. 
The task is to periodically estimate the location from the history of
sensor observations.
Formally, we seek to minimize the mean squared error, 
\begin{equation}\label{eq:cost}
\mathcal{L} = \sum_t{(\overline{x}_t-x^*_t)^2 + (\overline{y}_t-y^*_t)^2 + \beta (\overline{\phi}_t-\phi^*_t)^2 },
\end{equation}
where $\overline{x}_t,\overline{y}_t, \overline{\phi}_t$ and $x^*_t, y^*_t, \phi^*_t$ are the estimated
 and true robot poses for time $t$, respectively; $\beta$ is a
 constant parameter. 
 
Challenges are threefold. 
First, we must periodically update a posterior over
states given ambiguous observations, where the posterior is a 
multimodal, non-Gaussian, continuous distribution. PF-net tackles 
the challenge by encoding suitable differentiable algorithm prior, \ie, particle filtering.

Second, we need to build an observation model, $Z(o_t | s_t^k; \map)$,
that defines the probability of a camera observation, $o_t$, 
conditioned on the particle state, $s_t^k$, and the floor map, $\map$.   
A generative observation model would be hard to learn: it is 
a conditional distribution of images that show environments with different layouts, 
different furniture in different configurations, etc. PF-net learns a
discriminative model instead, which takes $s_t^k, o_t$ and $\map$ as inputs,
and the particle likelihood as output. The discriminative model only needs to learn features
relevant to localization. Importantly, PF-net learns the discriminative model end-to-end, 
without supervision on the particle likelihoods. 

Third, the observation model must compare geometry extracted
from a schematic \twod map and a camera image. 
This is especially hard for visually rich environments, where
some objects in the environment 
are not in the map, \eg, furniture.  
We introduce a neural network with appropriate structure
for the observation model~(\figref{fig:obsmodel}). 
The observation model defines the particle likelihood by appropriately combining 
features of the map and the camera image.
First, a \emph{local map} is obtained from $\map$ and $s^k_t$ through
an affine image transformation. For a neural network implementation we
adopt the Spatial Transformer Network~\cite{jaderberg2015spatial}. 
An affine transformation matrix, $A^k$, is computed for each particle, such 
that the transformed map is a local view from the pose, $s^k_t$. 
Because of this transformation, our network applies to any map size.
Next, we extract features both from the local map and the camera image through 
separate learned CNN components. The feature maps are concatenated, 
fed to a locally fully connected layer, reshaped to a vector, and fed to a 
final fully connected layer. 
The output is a scalar, $f^k_t$, the likelihood of the particle state.

The details of the PF-net are as follows.
Inputs are image observations, $o_t$, odometry, $u_t$, and the map, $\map$.
The output is the continuous pose, 
$\dense \overline{s_t} = \{\overline{x_t}, \overline{y_t}, \overline{\phi_t}\}.$
Particles are pairs of a candidate pose, $s_t^k$, and
weight, $w_t^k$. The output is the weighted mean of particles.
The loss for end-to-end training, $\mathcal{L}$, is defined by~\eqref{eq:cost}.
The observation and  transition models, $Z$ and $T$, are neural network components.
We discussed the observation model above.
The transition model updates the particle state given the odometry input, $u_t$, 
which is the relative motion defined in the robot coordinate frame.
Our neural network transforms $u_t$ to the global coordinate frame, $u_t'$, and 
adds it to the previous pose along with Gaussian noise. Formally, we sample the new
particle state from the differentiable function, 
$T(s_t^k | u_t', s_{t-1}^k) =  s_{t-1}^k + u_t' + \textrm{diag}(\sigma_t, \sigma_t, \sigma_r) \mathcal{N}(0; I)$,
where  $\mathcal{N}(0; I)$ is noise input from a standard multivariate Gaussian;
$\sigma_t$ and $\sigma_r$ are standard deviations for translation and rotation, respectively.
In experiments we chose $\sigma_t$ and $\sigma_r$ manually;
however, they could be learned  in the future. 

\section{Simulation experiments}

We implemented\footnote{
Our Tensorflow  implementation 
is available at \url{https://github.com/AdaCompNUS/pfnet}.}  and
evaluated PF-net in simulation for robot visual localization in indoor environments.
We compared with several  alternative methods to examine the benefits 
of differentiable algorithm priors and those of end-to-end training. 
We evaluated PF-net for various visual and depth sensor.  
Finally, we evaluated PF-net with increasing levels of uncertainty when the
robot's initial belief changes from a distribution concentrated around its
true pose to that of one spread uniformly over the entire space. 
The results are summarized in Table~\ref{tab:main}.

\begin{figure}[!t]
\begin{floatrow}
\ffigbox[.60\textwidth]{%
    \centering \hspace{-0.3cm}\includegraphics[height=2.4cm]{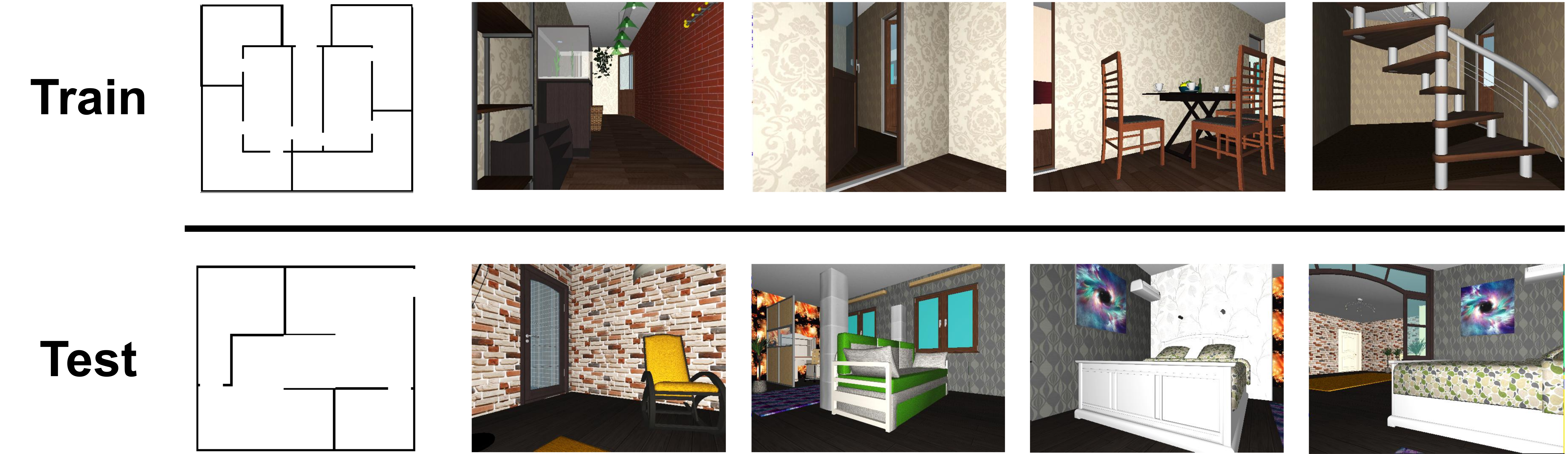}
}{\caption{Sample training and test data from the House3D data set. 
\vspace{-6pt}%
}\label{fig:examples}
}
\quad
\ffigbox[.34\textwidth]{%
\centering \hspace{-0.15cm}\includegraphics[height=2.5cm]{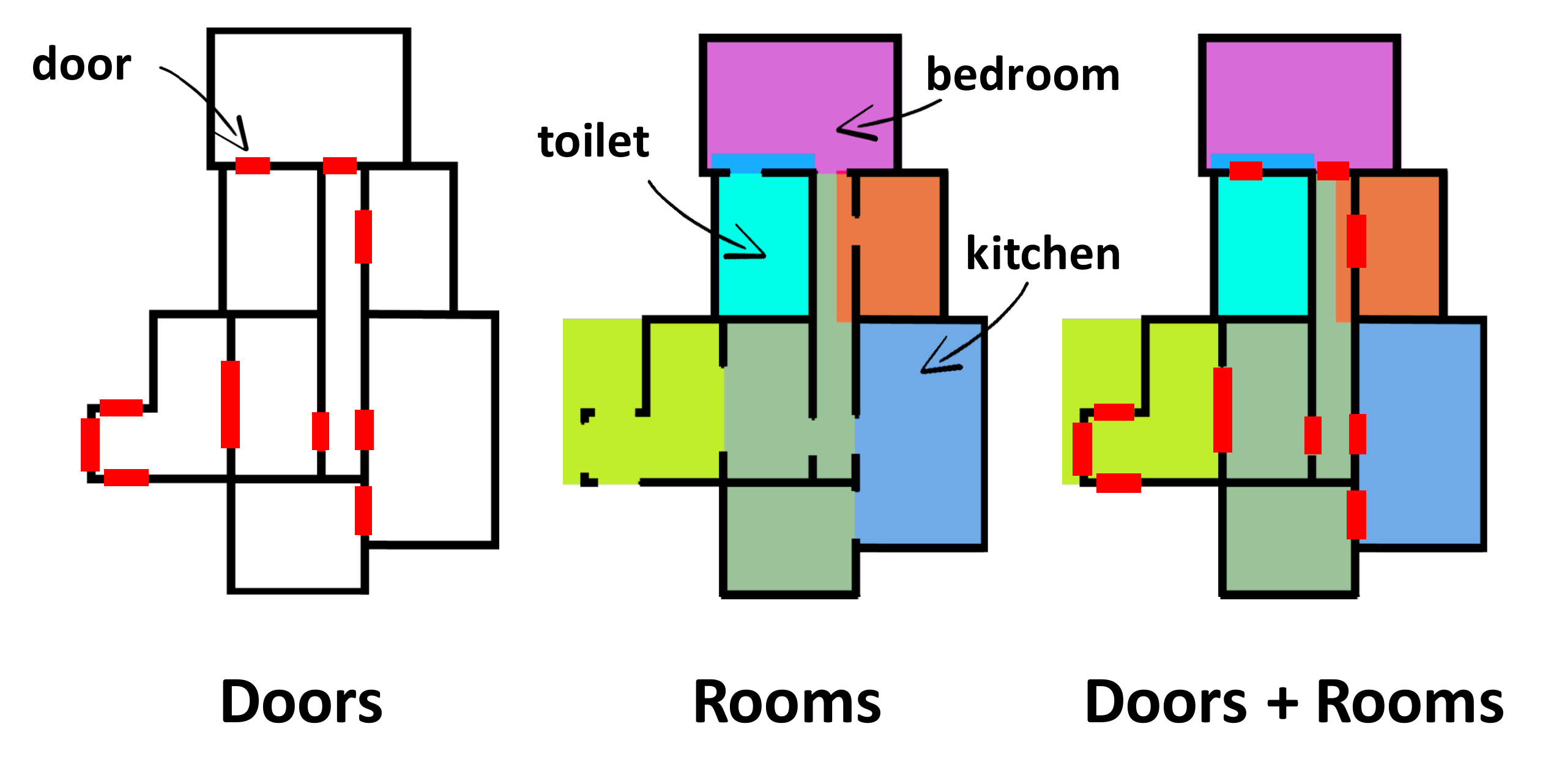}
}{\caption{Semantic maps labeled with doors and room types. 
\vspace{-6pt}%
} \label{fig:semantic_maps}
}
\end{floatrow}
\end{figure}

\subsection{Experimental setup}

\myparagraph{Simulation.} 
We conducted experiments in the House3D simulator~\cite{wu2018building}, 
which builds on a large collection of human-designed, realistic residential buildings 
from the SUNCG data set~\cite{song2017semantic}.
On average, the building 
size is $206\,\textrm{m}^2$, and the room size is $37\,\textrm{m}^2$.
See \figref{fig:examples} for examples. 

\myparagraph{Tasks.}
We consider localization with various levels of uncertainty. For
\emph{tracking}, the initial belief is concentrated around the true
state. For \emph{global localization}, the belief is uniform
over all rooms in a building. In between, for  \emph{semi-global
  localization}, the belief is uniform over 
 one or more rooms.

\myparagraph{Sensors.} 
We considered a monocular RGB camera, a depth camera, 
an RGB-D camera, and a simulated \twod LIDAR. 
Following earlier work~\cite{mendez2017sedar}, our simulated LIDAR
simply transforms a depth images to a \twod laser scan. The 
simulated LIDAR has a limited resolution of 54 beams and 
$\ang{60}$ field of view. As a result, localization with the simulated 
LIDAR is harder compared with a typical real-world LIDAR.
We also considered a simplified environment, LIDAR-W, for the LIDAR sensor 
by removing all furniture from the environment and leaving only the walls.
This way, the corresponding floor map contains all geometric objects in the
environment.

\myparagraph{Training.}  The training data consists of 45,000
trajectories from $200$ buildings.  Trajectories are generated  at
random:
the robot moves forward~($\dense p=0.8$) or turns~($\dense p=0.2$). 
The distance and the turning angle are sampled uniformly
from the ranges  $[20\,\textrm{cm}, 80\,\textrm{cm}]$
and $[\ang{15}, \ang{60}]$, respectively.
Each trajectory is 24 steps long, and each
step is labeled with the true robot pose. 
The
robot's initial belief $b_0$ is a multivariate Gaussian distribution.  The
center of $b_0$ is perturbed from the true pose  according to a Gaussian
with zero mean and covariance matrix
$\dense \Sigma=\textrm{diag}(30\,\textrm{cm}, 30\,\textrm{cm}, \ang{30})$, and the
the covariance of $b_0$ is the same $\Sigma$.  This setting corresponds to a
tracking task.
We trained PF-net and alternative networks
to minimize the end-to-end loss~\eqref{eq:cost}. 
We trained by backpropagation through time, limited to 4 time steps.
For training PF-net, we used $\dense K=30$ particles. We did not resample particles
during training, as it is not required for short trajectories and concentrated initial beliefs.

\myparagraph{Alternative methods.} 
We compared PF-net with alternative network architectures, 
histogram filter~(HF) network~\cite{jonschkowski2016} and 
LSTM network~\cite{hochreiter1997long}. 
HF network represents the belief as a histogram over discretized states, in
this case, a  grid with $40\,\text{cm} \times 40\,\text{cm} $ cells and $16$ orientations.  
Finer discretization did not produce better results. 
The LSTM network relies on its hidden state vector to
 represent the belief. We used a network architecture based on local
 maps, similar to the PF-net observation model. Outputs are
 relative state estimates that are updated with the odometry.
We also considered a conventional particle filtering (PF) method with a
handcrafted analytic observation model. We used the beam model implementation
from the AMCL package of ROS~\cite{quigley2009ros}, a standard model for 
localization with LIDAR~\cite{thrun2005probabilistic}.  The model
parameters were tuned for our simulated LIDAR sensor.
Finally, to calibrate the results, we also considered Odometry-NF, which
updates the belief only with odometry, not with other sensor inputs. 

\myparagraph{Evaluation.}  We evaluated the methods on a fixed set of 820
trajectories in 47 previously unseen buildings for tracking, semi-global
localization, and global localization tasks.  We used the same setup and same
model and algorithm parameters for all methods whenever possible.  We trained
the networks once for the tracking task and did not retrain for the other tasks.
It is important to observe that for PF-net, the number of particles, $K$, used in
execution does not have to be the same as that for training. In particular, we
used $\dense K=300$ particles for tracking, $\dense K=1,000$ for
semi-global localization, and $\dense K$ up to $\dense 3,000$ for global
localization.  We also activated resampling for semi-global and global
localization.  The same settings were applied to the PF method.
For tracking,    we report the average
root mean squared error~(RMSE), computed for the robot position
(Table~\ref{tab:main}a). 
For semi-global and global localization, we report success rate on 100-step
long trajectories (Table~\ref{tab:main}b--c). 
Localization is successful if the estimation error is below $1 \,\text{m}$ for the 
last 25 steps of a trajectory. Finally, we evaluated PF-net on semi-global
localization with semantic maps (Table~\ref{tab:main}d).

\subsection{Main results}
\begin{figure}[!t]
\begin{center}
\centerline{\includegraphics[width=0.95\textwidth]{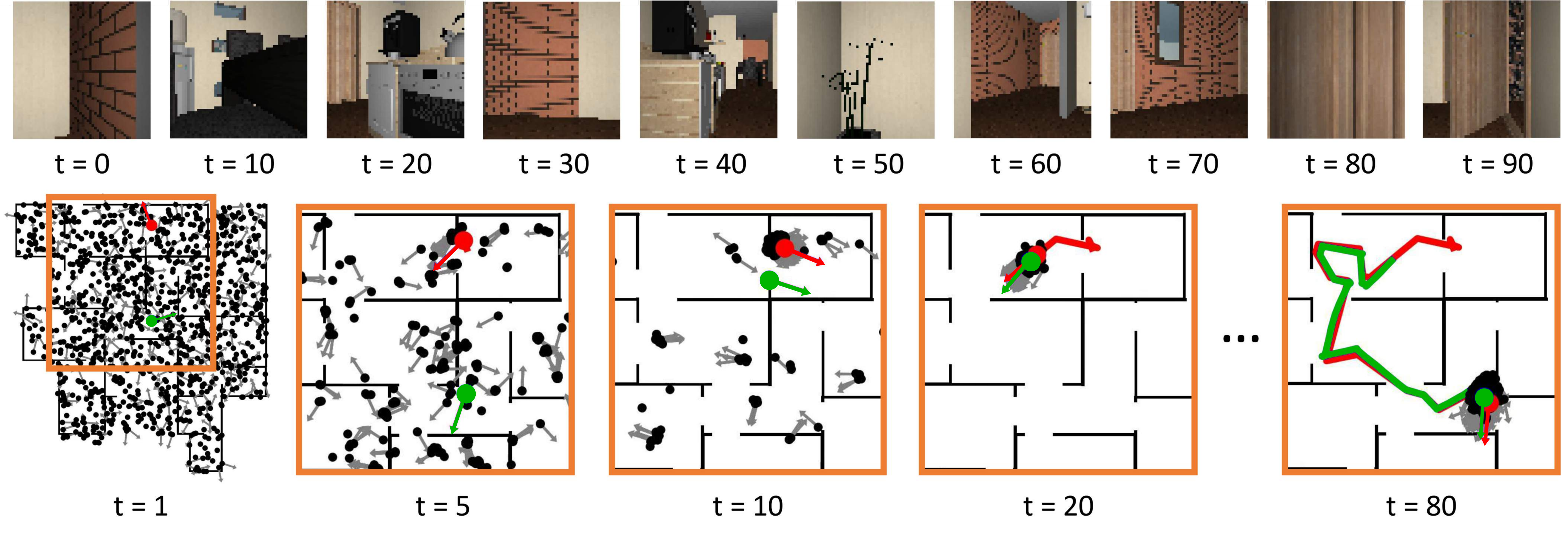}}
\caption{Global localization with PF-net ($K=1000$) and RGB camera input.
  Red indicates ground-truth poses. Black indicates PF-net particles.
  Green indicates the weighted mean of particles. 
\vspace{-6pt}
}
\label{fig:example_traj}
\end{center}
\vskip -0.3in
\end{figure}

PF-net successfully reduces state uncertainty in the tracking 
task~(Table~\ref{tab:main}a). Without additional training,
PF-net can also localize successfully when the initial belief is 
uniform over a room~(Table~\ref{tab:main}b),
and even when uniform over the entire floor map~(Table~\ref{tab:main}c).
See~\figref{fig:example_traj} for an example. 

\myparagraph{Differentiable algorithm priors are useful.} 
PF-net consistently outperformed alternative end-to-end learning architectures, 
HF network and LSTM network. Why? PF-net encodes 
differentiable algorithm prior specific to sequential state estimation, \ie, 
the particle representation of beliefs and their Bayesian update.
HF network encodes similar prior for updating beliefs, however,
it is restricted to a discrete belief representation which does not scale well to 
large and continuous state spaces.
The LSTM network is not restricted to a discrete state space, but it has no
structure specific to probabilistic state estimation, and it must
rely on the hidden state vector to encode the belief.

\begin{table}[!t]
\begin{floatrow}
\capbtabbox[0.6\textwidth]{
\scalebox{0.80}{
\begin{tabular}{@{\hskip 4pt}lccccc@{\hskip 4pt}}
  \toprule
 & RGB & Depth & RGB-D & LIDAR &      LIDAR-W \\
\midrule
PF-net  & ~~\textbf{40.5}   &   ~~\textbf{35.9}         &   ~~\textbf{33.3}    &   ~~\textbf{48.3}            &   ~~31.5      \\
HF network      &  ~~92.0         &    ~~91.6          &   ~~89.8      &  ~~95.6     &   ~~92.4     \\
LSTM network         & ~~66.9          &   ~~58.8          &   ~~60.3     &   ~~74.2      &   ~~64.4    \\
PF       &   --           &      --             &     --      &     ~~81.3        &  ~~\textbf{31.3}    \\
 Odometry-NF        &  109.4         & 109.4           & 109.4    &  109.4       & 109.4    \\
  \bottomrule 
\end{tabular}
}  
}{%
\vspace{-3pt}
  \caption*{\footnotesize (a) RMSE (cm) for tracking. 
	\vspace{-9pt}
  }
  \label{tab:tracking}
}
\quad
\capbtabbox[0.34\textwidth]{
\scalebox{0.80}{
\begin{tabular}{@{\hskip 4pt}lcccc@{\hskip 4pt}}
\toprule
$K$ & $N=1$ & $N=2$  & All  \\
\midrule
$500$	         &  80.0\pct    &  70.5\pct       & 46.1\pct     \\
$1,000$	     &  84.3\pct    & 80.1\pct       & 57.9\pct     \\
$2,000$	     &  87.3\pct     &  84.8\pct       & 68.5\pct     \\
$3,000$	      &  \textbf{89.0}\pct    & \textbf{85.9}\pct       & \textbf{76.3}\pct     \\
\bottomrule 
\end{tabular}
}}{%
\caption*{\footnotesize (c) PF-net success rate (\%) for semi-global and
  global localization, with $K$ particles.
  The initial belief is uniform over $N=1$, $N=2$, or all rooms.
  \vspace{-9pt}
  }
\label{tab:global}
}
\end{floatrow}
\end{table}


\begin{table}[!t]
\RawFloats
\begin{floatrow}
\capbtabbox[0.6\textwidth]{
\scalebox{0.80}{
\begin{tabular}{@{\hskip 4pt}lccccc@{\hskip 4pt}}
\toprule
 & RGB & Depth & RGB-D & LIDAR &      LIDAR-W \\
\midrule
PF-net  &    \textbf{82.6}\pct     &     \textbf{84.0}\pct       &  \textbf{84.3}\pct     &      \textbf{69.4}\pct         &   \textbf{86.6}\pct       \\
HF network      &  ~~2.7\pct         &    ~~2.9\pct        &   ~~4.5\pct           &  ~~1.6\pct       &  ~~2.2\pct     \\
LSTM network      &  21.1\pct     &    24.4\pct      &  23.4\pct     & 17.2\pct         &   22.2\pct    \\
PF     &   --           &      --             &     --      &     25.1\pct    &    86.2\pct    \\
Odometry-NF      &  ~~1.1\pct        &  ~~1.1\pct         &  ~~1.1\pct     &   ~~1.1\pct       &  ~~1.1\pct     \\
\bottomrule 
\end{tabular}
}  
}{%
 \vspace{-3pt}
  \caption*{\footnotesize (b) Success rate (\%) for semi-global localization
    over one room.
  \vspace{4pt}
  }
\label{tab:localization}
}
\quad
\capbtabbox[0.34\textwidth]{
\scalebox{0.80}{
\begin{tabular}{@{\hskip 4pt}lcccc@{\hskip 4pt}}
\toprule
Labels & RGB & Depth & RGB-D  \\
\midrule
None		   		     &  82.6\pct  	  &  84.0\pct    &  84.3\pct     \\
Doors      		   &  84.4\pct     &  83.9\pct     &  84.5\pct     \\
Rooms           &  84.5\pct    &  86.2\pct       &  86.5\pct     \\
Both              &  \textbf{84.6}\pct  &  \textbf{86.7}\pct        &  \textbf{87.2}\pct     \\
\bottomrule 
\end{tabular}
}}{%
\vspace{-3pt}
  \caption*{\footnotesize (d) PF-net success rate (\%) for semi-global
   localization with additional semantic information on  doors, rooms types, or both. 
  \vspace{-3pt}
  }
\label{tab:semantic}
}
\end{floatrow}
  \caption{Experimental comparison on robot visual localization.
   \vspace{-6pt}
  }\label{tab:main}
\end{table}

\myparagraph{End-to-end learning leads to increased robustness.}
We compared learned PF-net to PF with a known LIDAR 
model~(first and fourth row of Table~\ref{tab:main}a--b).
PF-net and PF performed similarly when only walls were present in the 
environment~(\emph{LIDAR-W} column).
PF-net performed significantly better when some objects in the
environment were not in the map~(\emph{LIDAR} column).
Why? The beam model has no principled way to distinguish 
relevant walls from irrelevant objects,
because it decouples the LIDAR scan to individual beams.
Through end-to-end training, PF-net may have learned relationships 
between beams to distinguish walls from objects.
PF-net may have also learned to deal with map imperfections, 
\eg{}, missing walls, glass doors, and various map artifacts,
 which we observed occasionally in the House3D data set. 

\myparagraph{PF-net is effective with various sensors.}
The columns of~Table~\ref{tab:main}a-b compare different sensors 
for localization. PF-net with RGB images is almost as effective as with depth
images; and it performs better than with simulated LIDAR.
This indicates that PF-net successfully learned to extract relevant geometry
from RGB images, and it learned to ignore objects that are not in the map. 
When combining RGB and depth image inputs, RGB-D column,
performance improves. This demonstrates that PF-net can learn simple sensor fusion
end-to-end from data. Future work in this direction is promising.

\subsection{Additional experiments}
\myparagraph{Global localization.}
We evaluated learned PF-net for localization with increasing difficulty~(Table~\ref{tab:main}c).
We chose initial beliefs uniform over one room, two rooms, and the entire building.
We compared PF-net with different number of particles, up to $\dense K=3000$. 
Results show that PF-net can solve global localization with high initial uncertainty when 
provided with sufficiently many particles.

\myparagraph{Semantic maps.}
Humans often use floor maps with semantic information: there are
labels for the office, toilet, lift and staircase.
Utilizing semantic maps for robot localization is not trivial~\cite{mendez2017sedar, mozos2006supervised}. 
PF-net may learn to use semantic maps naturally, through end-to-end training.
To demonstrate this, we trained PF-nets with simplified semantic maps
with labels for doors and room categories. See \figref{fig:semantic_maps} for examples.
We encoded the semantic labels in separate channels of the input map: 
one channel for doors, 8 channels for 8 distinct room categories.
Results show that simple semantic maps
can indeed improve localization performance~(Table~\ref{tab:main}d).

\myparagraph{Ablation study.} 
In supplementary experiments we altered certain settings of PF-net 
during training, and evaluated the learned PF-nets for a fixed
 semi-global localization task.
First, we added soft-resampling during training. 
When trained for the tracking task as before,
success rates decreased for soft-resampling: 79\% to 75\%. 
However, when trained with increased initial uncertainty and noisy odometry,
success rates increased for soft-resampling: 39\% to 42\%. 
As expected, resampling can be beneficial when most particles 
would be far from the true state; but it hurts when particles 
near the true state are eliminated, which often happens in 
early phases of learning.
Future work may incorporate various strategies for resampling only 
when required~\cite{doucet2009tutorial}.  
Indeed, when resampling only every second step, success rates increased: 42\% to 54\%.

Next, we varied the number of backpropagation steps for BPTT. 
Backpropagating through multiple steps improved performance: 
73\%, 79\%, 79\% success rates for 1, 2, and 4 
steps, respectively. This indicates that loss from future steps can 
provide a useful learning signal for the present step.

Finally, we replaced our loss function~\eqref{eq:cost}, with the 
probabilistic loss function proposed in~\cite{jonschkowski2018differentiable}. 
The alternative loss function worked worse when training
in the standard tracking setting, 74\% versus 79\% success rates. 
However, the alternative loss function worked better when 
training with increased uncertainty, 67\% versus 39\%. 
Our loss can be dominated by the distant particles, which 
may negatively affect learning in the latter case.

\section{Conclusion \& future work}
We introduced the PF-net, a neural network architecture with 
differentiable algorithm prior for sequential state estimation.
PF-net encodes learned probabilistic models, together with a particle filter algorithm, 
in a differentiable network representation. 
We applied PF-net to robot localization on a map.
 Through end-to-end training, 
PF-net successfully learned to localize in challenging, previously unseen 
environments populated with objects not shown in the map.

Future work may apply PF-net to real-world 
localization, a problem of great interest for mobile robot applications.
One concern is online execution. 
With RGB input PF-net
needs approx.~$0.6\textrm{ms}$ per particle per step. 
Indoor localization with high uncertainty may require up to 
1,000~--~10,000 particles~\cite{thrun2001robust}. 
We can increase robustness, and use less particles,  
by incorporating standard techniques for particle filtering, \eg,
injecting particles and adaptive resampling~\cite{doucet2009tutorial}.
We may also improve inference time, leveraging an abundance of work 
optimizing neural network models and hardware~\cite{sze2017efficient}.  
Finally, learned PF-net models can be used for standard 
particle filtering, and thus visual sensors can be 
complementary to laser, potentially at a lower update frequency.

PF-net could also be applied to other domains,
\eg{}, visual object tracking and SLAM. 
An exciting line of future work may extend PF-net to
learn latent state representations for filtering, potentially in an unsupervised setting. 
Finally, the particle representation of beliefs can be important for
encoding more sophisticated algorithms in neural networks,  
\eg{}, for planning under partial observability.

\clearpage

\acknowledgments{This research is supported in part by  Singapore Ministry of Education grant MOE2016-T2-2-068.
Peter Karkus is supported by the NUS Graduate School for Integrative Sciences and Engineering Scholarship.}


\small
\bibliography{references}  

\end{document}